\documentclass{article}

\usepackage{iclr2018_conference,times}

\usepackage{xcolor}
\usepackage{url}
\definecolor{darker}{rgb}{0,0.15,0.8}
\usepackage[colorlinks, urlcolor=darker, citecolor=darker, linkcolor=darker]{hyperref} 

\setcitestyle{round}

\usepackage{graphicx} 
\usepackage{subcaption} 

\usepackage{algorithm}
\usepackage{algorithmic}
\usepackage{hyperref}

\usepackage{booktabs}
\usepackage{relsize}
\usepackage{multirow}
\usepackage{rotating}
\usepackage{lipsum}
\usepackage{tabularx}
\usepackage{amsmath}
\usepackage{amsthm}
\usepackage{amssymb}
\usepackage{listings}

\usepackage{verbatim}

\newcommand{\x}[0]{$\times$}

\title{Block-Sparse Recurrent Neural Networks}

\author{
  Sharan Narang\\
  \texttt{sharan@baidu.com} \\
  Baidu Research \\
  \And
  Eric Undersander \\
  \texttt{undersandereric@baidu.com} \\
  Baidu Research
  \And
  Gregory Diamos \\
  \texttt{gregdiamos@baidu.com} \\
  Baidu Research \\
}

\begin{document}

\maketitle

\begin{abstract}

Recurrent Neural Networks (RNNs) are used in state-of-the-art models in domains such as speech recognition, machine translation, and language modelling. Sparsity is a technique to reduce compute and memory requirements of deep learning models. Sparse RNNs are easier to deploy on devices and high-end server processors. Even though sparse operations need less compute and memory relative to their dense counterparts, the speed-up observed by using sparse operations is less than expected on different hardware platforms. In order to address this issue, we investigate two different approaches to induce \emph{block} sparsity in RNNs: pruning blocks of weights in a layer and using group lasso regularization to create blocks of weights with zeros. Using these techniques, we demonstrate that we can create block-sparse RNNs with sparsity ranging from 80\% to 90\% with small loss in accuracy. This allows us to reduce the model size by roughly 10\x. Additionally, we can prune a larger dense network to recover this loss in accuracy while maintaining high block sparsity and reducing the overall parameter count. Our technique works with a variety of block sizes up to 32\x32. Block-sparse RNNs eliminate overheads related to data storage and irregular memory accesses while increasing hardware efficiency compared to unstructured sparsity.

\end{abstract}

\section{Introduction}
\label{sec:introduction}

Improvements in several applications such as speech recognition \citep{amodei2016deep}, language modeling \citep{DBLP:journals/corr/JozefowiczVSSW16}, and machine translation \citep{DBLP:journals/corr/WuSCLNMKCGMKSJL16} are a result of large Recurrent Neural Networks (RNNs) trained on large scale datasets. As the datasets available to train these models have grown, so have model sizes. Deployment of such large models is compute and memory intensive.

Pruning deep neural networks is an effective strategy to reduce the overall memory and compute requirements of these models \citep{narang2017exploring, han2015deep}. However, these approaches induce random, unstructured sparsity in the weight matrices. Speed-up obtained with random sparsity on various hardware platforms are lower than expected (as shown in \citet{narang2017exploring,deepbench}). Sparse formats do not efficiently utilize the hardware resources due to storage overheads, irregular memory access, and inability to take advantage of array data-paths in modern processors. 

Block sparsity can address these issues. Saving indices of non-zero blocks instead of indices for non-zero elements reduces the storage overhead by a factor of block size. Block-sparse formats store blocks contiguously in memory reducing irregular memory accesses. Block sparsity inherently allows us to take advantage of array-data-path in modern processors.


In order to induce block sparsity in RNNs, we propose a block pruning approach that zeros out blocks of weights in the matrix while the network is training. At the end of training, the algorithm creates a block-sparse RNN. In addition to this pruning technique, we examine the efficacy of group lasso regularization to induce block sparsity in the network. We also combine group lasso regularization with block pruning. 


We demonstrate that block pruning and group lasso regularization with pruning are successful in creating block-sparse RNNs. Inducing block sparsity with 4\x4 blocks in vanilla RNNs and Gated Recurrent Units (GRUs) \citep{cho2014learning} results in 9\% to 17\% loss in accuracy compared to the dense baseline. Model size reduces by nearly 10\x. Block sizes can be scaled up to 32\x32 with our approach. Larger blocks require lower sparsity to maintain similar accuracy. We can also reduce accuracy loss by starting with a larger dense matrix than the baseline and then pruning it down while still reducing the number of parameters compared to the baseline.

Our approach is agnostic to the optimization algorithm and does not require any hyper-parameter retuning (besides pruning and regularization hyper-parameters). Furthermore, since our approach does not require re-training the model, training time remains the same.

\section{Related Work}
\label{sec:related_work}

There have been several approaches to reduce the network size by pruning the model. \citet{Hanson:1989:CBM:89851.89872} use several bias techniques to decay weights in a network. \citet{Cun90optimalbrain} and \citet{hassibi1993optimal} both use Hessian-based approaches to prune weights below a certain threshold. 
Simpler approaches like sorting or thresholding can be used to prune a neural network. \citet{han2015deep} and \citet{liu2015sparse} prune Convolution Neural Networks (CNNs) while maintaining high accuracy.
\citet{yu2012exploiting} use a hard threshold to prune deep learning models. \citet{narang2017exploring} and \citet{zhu2017prune} prune recurrent neural networks during the initial training run with a small accuracy loss using gradual pruning. Unlike our technique, all of the above approaches induce random, unstructured sparsity in neural networks. 

Several approaches exist to induce structured sparsity in neural networks. \citet{mao2017sparsity} use a simple threshold based technique to create structurally sparse CNNs. \citet{yu2017scalpel} propose Scalpel that prunes CNNs taking into account the underlying target hardware architecture.   \citet{wen2017learning} alter the structure of Long Short Term Memory (LSTM) \citep{lstm} to create LSTMs with smaller memory footprint. They demonstrate that this technique works for language modeling on the Penn Tree Bank dataset. Our approach works with both vanilla RNN and GRU models trained on a large-scale datasets for speech recognition.

Group lasso regularization has been used as an efficient method for generating sparse structures \citep{yuan2006model, kim2010tree}. \citet{wen2016learning} use group lasso regularization to induce structured sparsity in convolutional neural networks. Regularization is a known method to induce sparsity in deep neural networks \citep{faraone2017compressing, fan2016}. To the best of our knowledge, none of these approaches have been used with RNNs trained on large-scale datasets.

Other approaches to reduce compute and memory footprint for deep learning models include quantization \citep{Micikevicius2017mixed, 37631, Rastegari2016,gupta2015deep} and low-rank factorization \citep{DBLP:journals/corr/DenilSDRF13, DBLP:journals/corr/DentonZBLF14}. Our approach is orthogonal to these methods and can be combined with them.

\section{implementation}
\label{sec:implementation}
\subsection{Block Pruning}
\label{sec:block_pruning}
Our approach to pruning deep learning models builds on the work by \citet{narang2017exploring}. They propose a weight pruning algorithm that introduces random, unstructured sparsity in RNNs. In their work, they propose pruning weights below a monotonically increasing threshold. Their pruning strategy does not impose any structure on the weights. 

\begin{figure}
\centering
\includegraphics[width=0.95\linewidth]{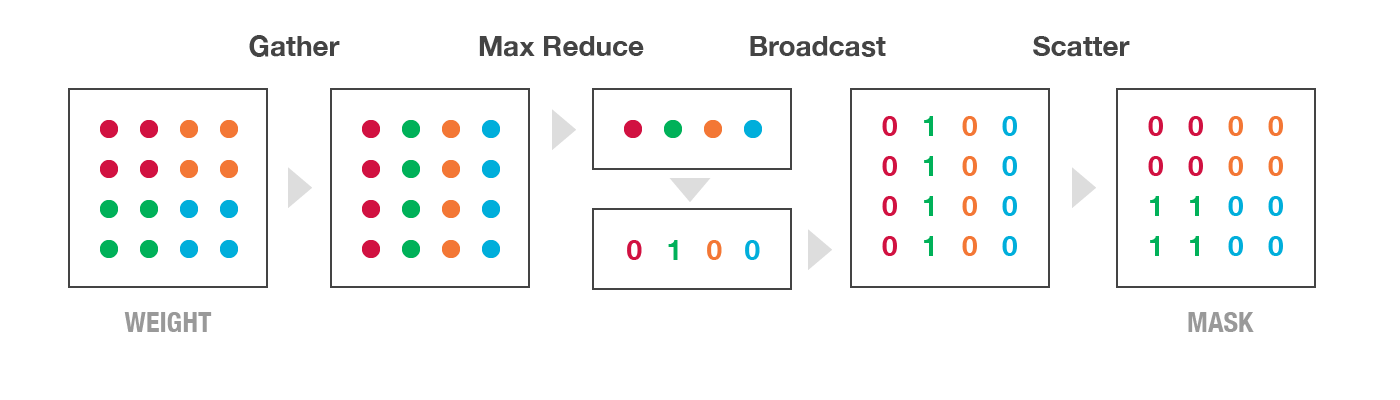}
\caption{Generating block-sparse masks from a weight matrix}
\label{fig:block_prune}
\end{figure}

We extend this approach to prune blocks of a matrix instead of individual weights. In order to prune blocks, we pick the weight with the maximum magnitude as a representative for the entire block. If the maximum magnitude of a block is below the current threshold, we set all the weights in that block to zeros. Figure \ref{fig:block_prune} depicts the process of generating a block-sparse mask from a weight matrix for a given threshold. The block-sparse mask is multiplied with the weights to generate block-sparse weight matrix. The monotonically growing threshold ($\epsilon$) causes more blocks to be pruned as training progress. We stop pruning more blocks after around 40\% of training has completed. Any blocks that had been zeroed out are held at zero even after pruning has ended resulting in a sparse model at the end of training. 


\citet{narang2017exploring} use six hyper-parameters to determine the threshold at a given iteration. Table~\ref{tab:hyper-param} provides the description and heuristics (adapted for block pruning) for these hyper-parameters. The \emph{start\_slope} and \emph{ramp\_slope} determine the rate at which the threshold increases. In order to determine \emph{start\_slope}, they recommend using weights from an existing dense model. To achieve 90\% sparsity, they assign \emph{q} to the weight which is the 90th percentile of the absolute values in a weight matrix. Assuming \emph{$\phi$} is 1.5\emph{$\theta$}, they use Equation \ref{eq:start_slope} to determine \emph{$\theta$}.



\begin{table}
\caption{Heuristics to pick hyper-parameters for block-pruning}
\label{tab:hyper-param}
\begin{center}
 \begin{tabular}{p{1cm}p{5cm}p{4cm}}
\multicolumn{1}{c}{\bf HYPER-PARAM}  &\multicolumn{1}{c}{\bf DESCRIPTION} &\multicolumn{1}{c}{\bf HEURISTIC VALUES}
\\ \hline \\
 \emph{start\_itr} & Iteration to start pruning & Start of second epoch\\  
 \emph{ramp\_itr} & Iteration to increase the rate of pruning & Start of 20\% of total epochs\\ 
 \emph{end\_itr} & Iteration to stop pruning more parameters & Start of 40\% of total epochs \\ 
 \emph{start\_slope} ($\theta$) & Initial rate of increasing the threshold & See Equation \ref{eq:start_slope_block} \\ 
 \emph{ramp\_slope} ($\phi$) & Rate of increasing threshold after ramp iteration & $1.2 \theta$ to $1.7 \theta$ \\ 
  \emph{freq} & Number of iterations after which $\epsilon$ is updated & 100 \\
\hline
\end{tabular} 
\end{center}
\end{table}


\begin{equation} \label{eq:start_slope}
    \theta = \frac{2 \times q \times freq}  {2 \times (\mathit{ramp\_itr} - \mathit{start\_itr}) + 3\times(\mathit{end\_itr} - \mathit{ramp\_itr})}
\end{equation}


For block pruning, we need to modify the \emph{start\_slope} to take into account the number of elements in a block (\emph{$N_b$}). In order to calculate the \emph{start\_slope}, we first calculate \emph{start\_slope} for weight pruning (\emph{$\theta_w$}) using the Equation \ref{eq:start_slope}. Given \emph{$\theta_w$}, we suggest using Equation \ref{eq:start_slope_block} to determine the initial slope (\emph{$\theta_b$}) for block pruning. Based on empirical results, we have found that using this approach allows us to achieve block sparsity ranging from 85\% to 95\%. Further tuning of these hyper-parameters is required to achieve desired block sparsity.

\begin{equation} \label{eq:start_slope_block}
    \theta_b = \theta_w \times \sqrt[4]{N_b}
\end{equation}

We prune all the recurrent and fully connected layers in the network using the same block size. The pruning hyper-parameters are same for each type of layer in the network - recurrent weight layer and linear/fully connected layer. 

\subsection{Group Lasso Regularization}
\label{sec:group_lasso}


Group lasso is a type of weight regularization that works on groups of weights and can zero out all the weights in a group. In order to induce block sparsity in the network, we divide all weights in the model into blocks. For each block, we add a loss term proportional to the $\ell_2$ norm of the block.
$$
L = L_{\text{training}} + \lambda_g \sum_{g=1}^G \|w^{(g)}\|_2
$$
where $w^{(g)}$ is a block of weights, $\|w^{(g)}\|_2$ is the $\ell_2$ norm of the block, and $G$ is the total number of block. Our use of $\ell_2$ norm is a variant of the more general group lasso defined in \cite{yuan} as $\|n\|_K = (n'Kn)^{1/2}$.

Group lasso has the property that a large enough $\lambda_g$ will drive all weights within certain groups to hard zeros. Thus, we explore group lasso regularization to produce block-structured sparsity. We choose an appropriate constant $\lambda_g$ for the duration of training. 

One interpretation of weight regularization is that less important weights are driven towards zero and more important weights retain large absolute values. Thus, we combine group lasso with block pruning, such that group lasso guides the selection of blocks to prune. We apply group lasso regularization to coincide with the pruning schedule. We turn off regularization when the pruning schedule ends, which is typically after around 40\% of training epochs. As discussed in Section \ref{sec:block_pruning}, weights that were already set to zero remain unchanged after this point. Group lasso is related to the well-known $\ell_1$ regularization. In Appendix \ref{appendix}, we discuss exploration of $\ell_1$ regularization combined with weight pruning.



\section{Experiments}
\label{sec:experiments}

We run block sparsity experiments on two different speech recognition models from \citet{amodei2016deep}. The RNN model consists of a convolutional layer, followed by seven bidirectional recurrent layers and a Connectionist Temporal Classification (CTC) layer \citep{graves2006connectionist}. The baseline RNN model (RNN Dense 1760) consists of 1760 hidden units in each recurrent layer with nearly 67 million parameters. The GRU model consists of two convolutional layers, three recurrent layers with GRU cells and a CTC layer. The baseline GRU model (GRU Dense 2560) consists of 2560 hidden units in each layer with a total of 115 million parameters. The dataset used for training these models consists of 2100 hours of English speech. We use a validation set consisting of 3.46 hours of data. The Character Error Rate (CER) results are reported on an independent test set, consisting of 2.9 hours of English data. 

In order to introduce block sparsity in RNNs, we run three different types of experiments - Block Pruning (BP), Group Lasso (GL), and Group Lasso with block pruning (GLP). We prune weights in the recurrent layers (both linear and recurrent weights) and fully connected layers. Biases, batch-normalization parameters and weights in the convolutional and CTC layers are not pruned since they account for a small portion of the total weights in the network. Besides pruning hyper-parameters and $\lambda_g$, no other hyper-parameter changes were required for sparse training runs. The models are trained using Nesterov Stochastic Gradient Descent (SGD) with momentum. All models are trained for 25 epochs. The dense models are trained without any regularization.

In Section \ref{sec:block_sparsity_results}, we report results for different sparse models pruned with 4\x4 blocks. Section \ref{sec:group_lasso_variants} compares the results for the two different group lasso experiments. Section \ref{sec:block_size} discusses the impact of varying the block size on the accuracy of the model.

\subsection{Block Sparsity}
\label{sec:block_sparsity_results}
We conduct three types of experiments for both RNN and GRU models: pruning the baseline model, training smaller dense models, and pruning a model larger than the baseline model. 

Initially, we prune the baseline RNN and GRU models. Using BP and GLP, we are able to reduce the parameter count for both these models by nearly 10\x. As shown in Table \ref{tab:results}, the sparse RNN model with 1760 hidden units has an overall block sparsity of 89\% with a relative loss in accuracy of 16.7\%. The sparse GRU model achieves slightly higher sparsity (90\%) while losing only 8.8\% of accuracy. This indicates that the block-sparse GRU model retains most of the capacity of the dense model.


\begin{table}
\caption{GRU and bidirectional RNN model results with 4\x4 blocks}
\label{tab:results}
\begin{center}
 \begin{tabular}{lrrrrl}
\multicolumn{1}{c}{\bf}
&\multicolumn{1}{c}{\bf \# PARAMS}
&\multicolumn{1}{c}{} 
&\multicolumn{1}{c}{} 
&\multicolumn{1}{c}{\bf RELATIVE} 
&\multicolumn{1}{c}{\bf PRUNING} 
\\
\multicolumn{1}{c}{\bf MODEL}
&\multicolumn{1}{c}{(in millions)}
&\multicolumn{1}{c}{\bf SPARSITY}
&\multicolumn{1}{c}{\bf CER} 
&\multicolumn{1}{c}{\bf PERF}
&\multicolumn{1}{c}{\bf ALGORITHM} 
\\ \hline \\
RNN Dense 1760 & 67 & 0.0\% & 15.36 & 0.0\%  & N/A \\
RNN Dense 704 & 11.6 & 0.0\% & 18.95 &  -23.4\% & N/A \\
RNN Sparse 1760 & 7.3 & 89.2\% & \bf 17.93 &  -16.7\%  & BP\\
RNN Sparse 2560 &  12.9 & 90.8\% &15.89 & -3.4\% & GLP\\
RNN Sparse 3072 &  25.8  & 87.3\% & \bf 15.66 & -1.9\% & BP\\ 
\hline \\
GRU Dense 2560 &  115  & 0.0\%  &15.42 & 0.0\% & N/A\\
GRU Dense 704 &  11.0  & 0.0\% &21.26 & -37.9\% & N/A \\
GRU Sparse 2560 &  10.8  & 90.6\% & \bf 16.78 & -8.8\% & GLP\\
GRU Sparse 3584 &  25.6 & 88.4\% & \bf 16.23 & -5.2\% & BP\\
\hline
\end{tabular}
\end{center}
\end{table}

Secondly, we train dense models with fewer parameters to determine if sparsity is reducing overfitting in the large dense baseline models. For both RNN and GRU models, we train a dense model with 704 hidden units in each layer, resulting in approximately the same number of parameters as the final sparse models. Table~\ref{tab:results} shows that these dense models perform worse than the sparse models for both RNN and GRU models. Large sparse models are a better approach to reduce parameter count than dense small models.




Finally, we train sparse models with more hidden units in each recurrent layers to recover the accuracy. For RNN models, we increase the hidden layer size to 2560 and 3072. As shown in Table \ref{tab:results}, the RNN sparse 3072 is only 1.9\% worse than the dense baseline model. The 2560 and 3072 sparse RNN models reduce the overall parameter count by 5\x and 2.5\x respectively. Similarly, pruning the GRU model with 3584 hidden nodes reduces the accuracy loss to about 5\% while still shrinking the model by 4.5\x.




Our evaluation show that inducing block sparsity in the baseline model allows us to reduce the model size by approximately 10\x with a small loss in accuracy. Pruning a model larger than the baseline model allows to reduce the accuracy loss while reducing model size by nearly 5\x. Our results also indicate that large sparse models result in better accuracy that small dense models. 

\subsection{Group Lasso Variants}
\label{sec:group_lasso_variants}

Table \ref{tab:gl-results} highlights the results of GL and GLP experiments for two different models. For both RNN models with 1760 and 2560 hidden nodes, group lasso without any pruning does significantly worse than combining group lasso with the block pruning methodology.

\begin{table}[h]
\caption{Group lasso experiments for RNN models with 4\x4 blocks}
\label{tab:gl-results}
\begin{center}
 \begin{tabular}{lrrrrl}
\multicolumn{1}{c}{} 
&\multicolumn{1}{c}{\bf \# PARAMS}
&\multicolumn{1}{c}{}
&\multicolumn{1}{c}{} 
&\multicolumn{1}{c}{\bf RELATIVE} 
&\multicolumn{1}{c}{\bf PRUNING} 
\\
\multicolumn{1}{c}{\bf MODEL} 
&\multicolumn{1}{c}{(in millions)} 
&\multicolumn{1}{c}{\bf SPARSITY}
&\multicolumn{1}{c}{\bf CER}
&\multicolumn{1}{c}{\bf PERF} 
&\multicolumn{1}{c}{\bf ALGORITHM} 
\\ \hline \\
RNN Sparse 1760 & 10.9  & 83.3\% &30.14 &  -96\%  & GL \\
RNN Sparse 1760  & 6.2  & 90.8\% & \bf 19.24& -25.3\%  & GLP \\
RNN Sparse 2560  & 24.4  & 82.8\% & 27.4& -78.4\% & GL\\
RNN Sparse 2560  & 12.9  & 90.8\% & \bf 15.89 & -3.4\% & GLP\\
\hline
\end{tabular}
\end{center}
\end{table}

In order to achieve high sparsity (80\% or higher), we need to set $\lambda_g$ to a relatively high value. For instance, experiments using GL required a $\lambda_g$ of approximately 3\x  larger than the GLP experiments. This high regularization factor hurts the model accuracy. The dense baseline model is trained without any regularization. Even without regularization, the dense model does not overfit the training dataset. Group lasso experiments underfit the training data due to the high value of $\lambda_g$. Group lasso could be more successful in inducing sparsity where the dense model overfits the training dataset. In the GLP experiments, we can reduce the regularization factor since pruning forces smaller magnitude weights to zero. This combined approach results in improved accuracy while maintaining high levels of sparsity. 

\subsection{Block Size Variation}
\label{sec:block_size}

Table~\ref{tab:blocks-results} shows the results of varying block size for pruning for RNN and GRU baseline models. Increasing the block size to 16\x16 and 32\x32 requires reducing the sparsity to 83.6\% and 79.1\% respectively for RNN models to obtain good accuracy. Similar results hold true for the GRU model as well. Large sparse blocks reduce memory overhead for storing non zero values and can take advantage of array data-paths in more modern processors. Therefore, even though large blocks achieve lower sparsity, they result in lower memory and compute requirements. 



\begin{table}[t]
\caption{GRU and bidirectional RNN results for different block sizes using BP}
\label{tab:blocks-results}
\begin{center}
\begin{tabular}{llrrrr}
\multicolumn{1}{c}{}
&\multicolumn{1}{c}{\bf BLOCK} 
&\multicolumn{1}{c}{\bf \# PARAMS}
&\multicolumn{1}{c}{} 
&\multicolumn{1}{c}{} 
&\multicolumn{1}{c}{\bf RELATIVE} 
\\
\multicolumn{1}{c}{\bf MODEL}
&\multicolumn{1}{c}{\bf SIZE} 
&\multicolumn{1}{c}{(in millions)}
&\multicolumn{1}{c}{\bf SPARSITY}
&\multicolumn{1}{c}{\bf CER}
&\multicolumn{1}{c}{\bf PERF} 
\\ \hline \\
RNN Sparse & 1x1 & 7.3 &  89.2\% & 17.32 & -12.8\%  \\
RNN Sparse & 4x4 &  7.3 & 89.2\% & 17.93 & -16.7\%   \\
RNN Sparse & 12x2 & 10.8 & 84.1\% & 16.96 & -9.99\%   \\
RNN Sparse & 8x8  & 10.7 & 84.1\% & 17.66 & -14.9\%   \\
RNN Sparse & 16x16 &  11.1 & 83.6\% & 17.1 & -11.3\%  \\
RNN Sparse & 32x32 &  14.1 & 79.1\% & 16.67 & -8.5\%  \\
\hline
GRU Sparse & 1x1 &  13.1 & 88.7\% & 16.55 & -7.3\%   \\
GRU Sparse & 4x4 &  16.2 &  86.0\% &  16.97 & -10.5\% \\
GRU Sparse & 16x16 &  20.8 & 81.9\% & 16.84 & -9.2\%  \\
\hline
\end{tabular}
\end{center}
\end{table}

\section{Performance}
\label{sec:perf}

The primary advantage of a block-sparse format is to increase hardware efficiency by making the computation more regular. Sparse formats incur at least three types of overhead: i) indexing overhead, ii) irregular memory accesses, and ii) incompatibility with array-data-paths, all of which are mitigated by using larger block sizes.  

\begin{figure}
\begin{subfigure}{0.49\linewidth}
\centering
\includegraphics[width=0.95\linewidth]{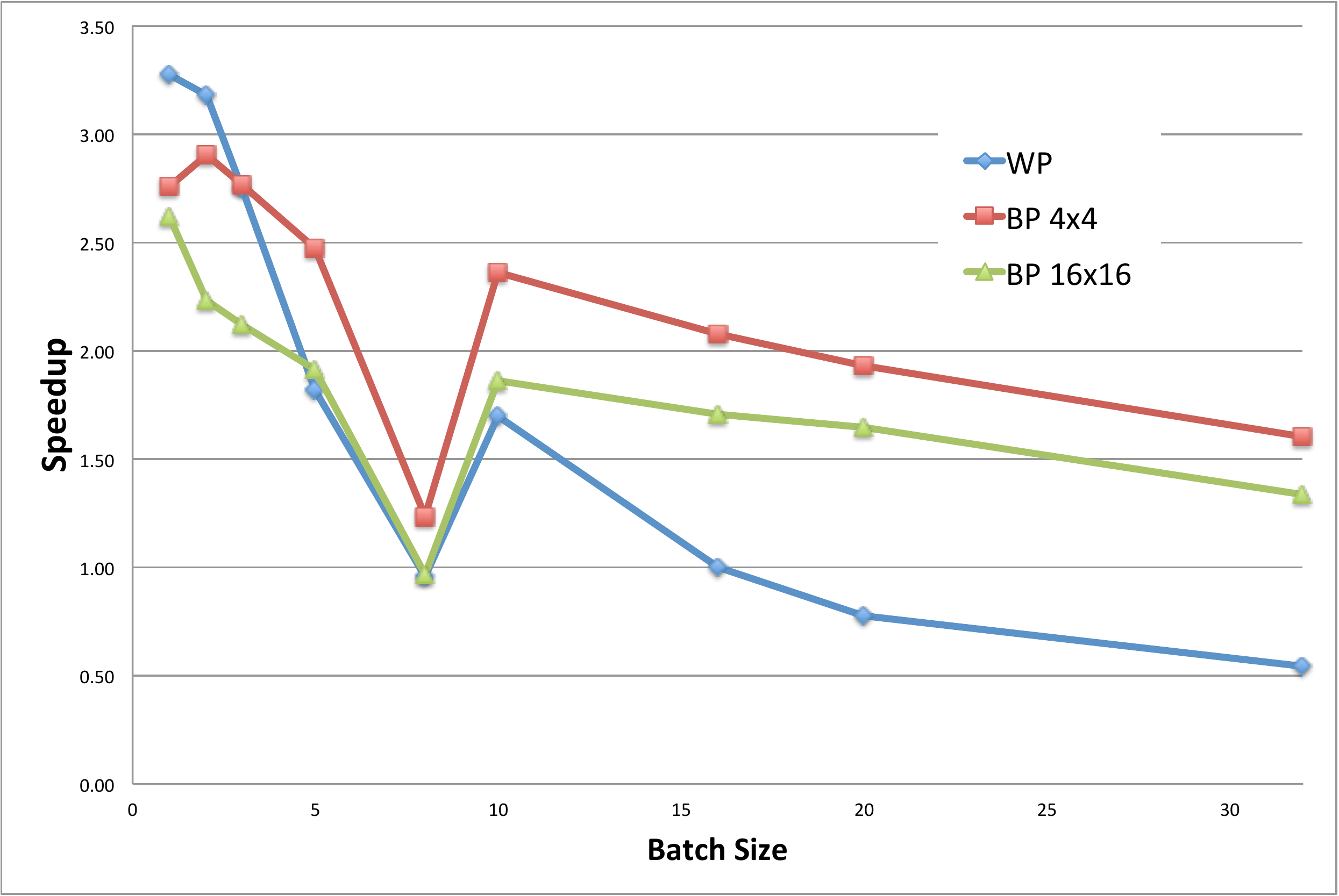} 
\caption{Speed-up for RNN 1760 layer matrix multiply}
\label{fig:rnn_speedup}
\end{subfigure}
\begin{subfigure}{0.49\linewidth}
\includegraphics[width=0.95\linewidth]{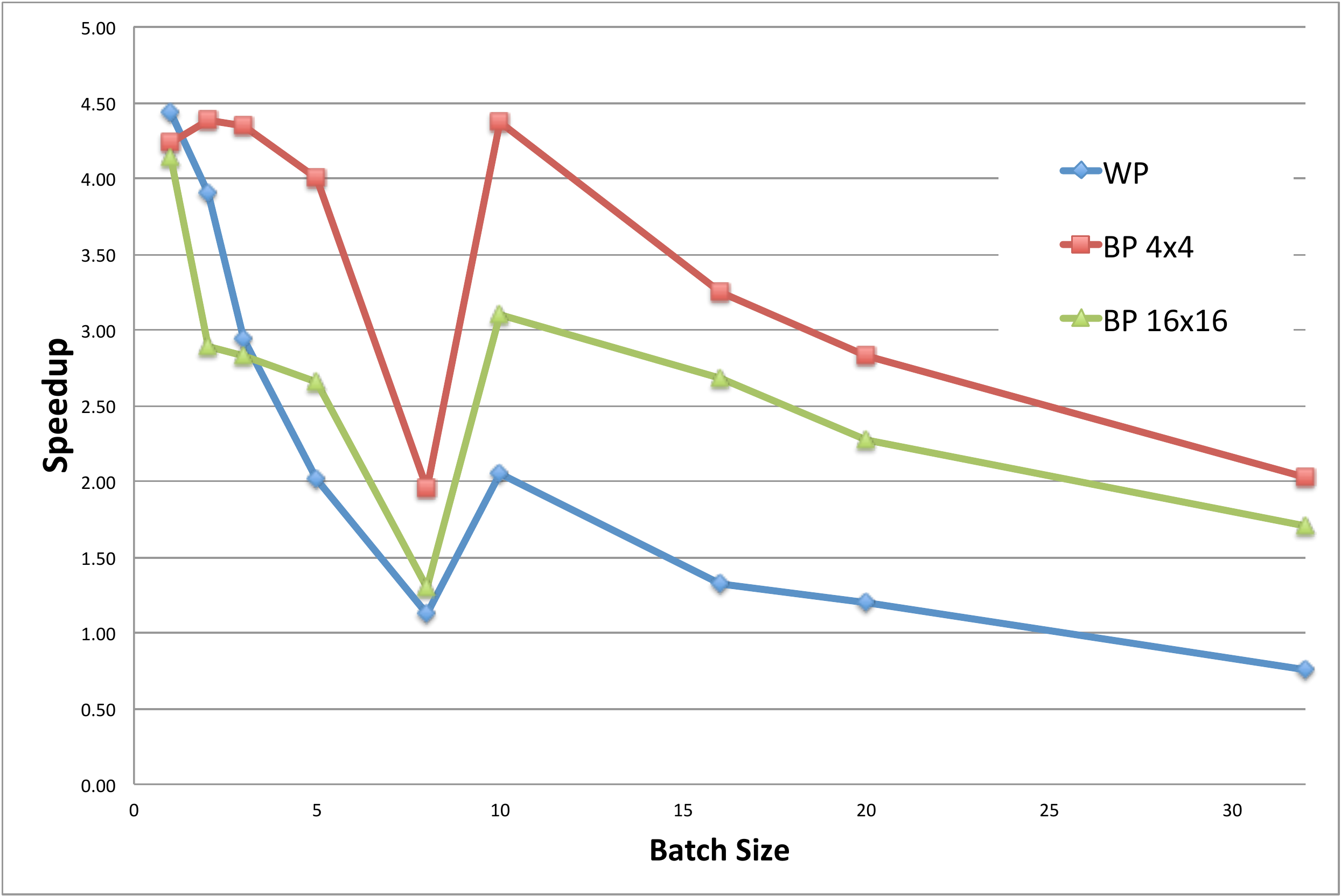}
\caption{Speed-up for GRU 2560 layer matrix multiply}
\label{fig:gru_speedup}
\end{subfigure}
\caption{Speed-up for sparse matrix dense matrix multiply. Benchmarks are run on TitanX Maxwell using the CuSparse library. Sparse matrices are represented in the CSR format. RNN matrix sizes are (1760,1760) with 90\% sparsity and (1760, batch\_size). GRU matrix sizes are (7680,2560) with 95\% sparsity and (2560, batch\_size). Results are shown for matrices from Weight Pruning~(WP) and Block Pruning~(BP).}
\label{fig:speedup}
\end{figure}

\textbf{Indexing Overheads}.
Sparse formats use extra memory to track the location of each non-zero value.  For example, the compressed-sparse-row (CSR) format uses approximately two extra index values for each non-zero value.  The size of these extra index values depends on the maximum matrix size.  Using 16-bit indices incurs 32-bits of overhead per non-zero value and allows up to 64k x 64k matrices to be supported.  Assuming that neural network weights are represented with 16-bits as in~\cite{Micikevicius2017mixed}, this is a $200\%$ overhead.  Block sparsity reduces this overhead by a factor of the block size because the index is shared over the entire block.  For example, using a block size of 4x4 reduces the memory bloat to $12.5\%$, and using a block size of 16x16 reduces the overhead to less than $1\%$. 

\textbf{Irregular Memory Accesses}.
Caches lines, DRAM row buffers, and TLBs provide the best performance when memory is accessed in relatively large contiguous units (e.g. 64 bytes for cache lines, 4KB for a DRAM row) as opposed to in fine-grained random accesses.  Block-sparse formats store blocks contiguously in memory, resulting in large coalesced accesses. 

\textbf{Array Data-Paths}.
Fine-grained sparsity cannot directly take advantage of array-data-paths, such as the 16x16 TensorCore units in the Volta GPU described by \cite{volta2017} or the 256\x256 units in the Google TPU described by \cite{Jouppi2017tpu}.  There are significant advantages of using these units, for example, on the Volta V100 GPU, they enable up to 8x higher throughput than the SIMD data-paths. In order to keep these units busy, the block size should be at least as large as the hardware data-path size (i.e. 16\x16 or greater on V100).

Figure \ref{fig:speedup} shows that block-sparse matrices achieve higher speed-up than unstructured sparsity for large batch sizes. In this case, the speed-up is achieved due to reducing irregular memory accesses and improving load balance. 4\x4 blocks have higher speed-up than 16\x16 blocks. Further investigation is needed to understand this behavior. 







\section{Discussion}
\label{sec:discussion}



\subsection{Pruning Characteristics}

\begin{figure}
\begin{subfigure}{0.49\linewidth}
\centering
\includegraphics[width=0.95\linewidth, height=5cm]{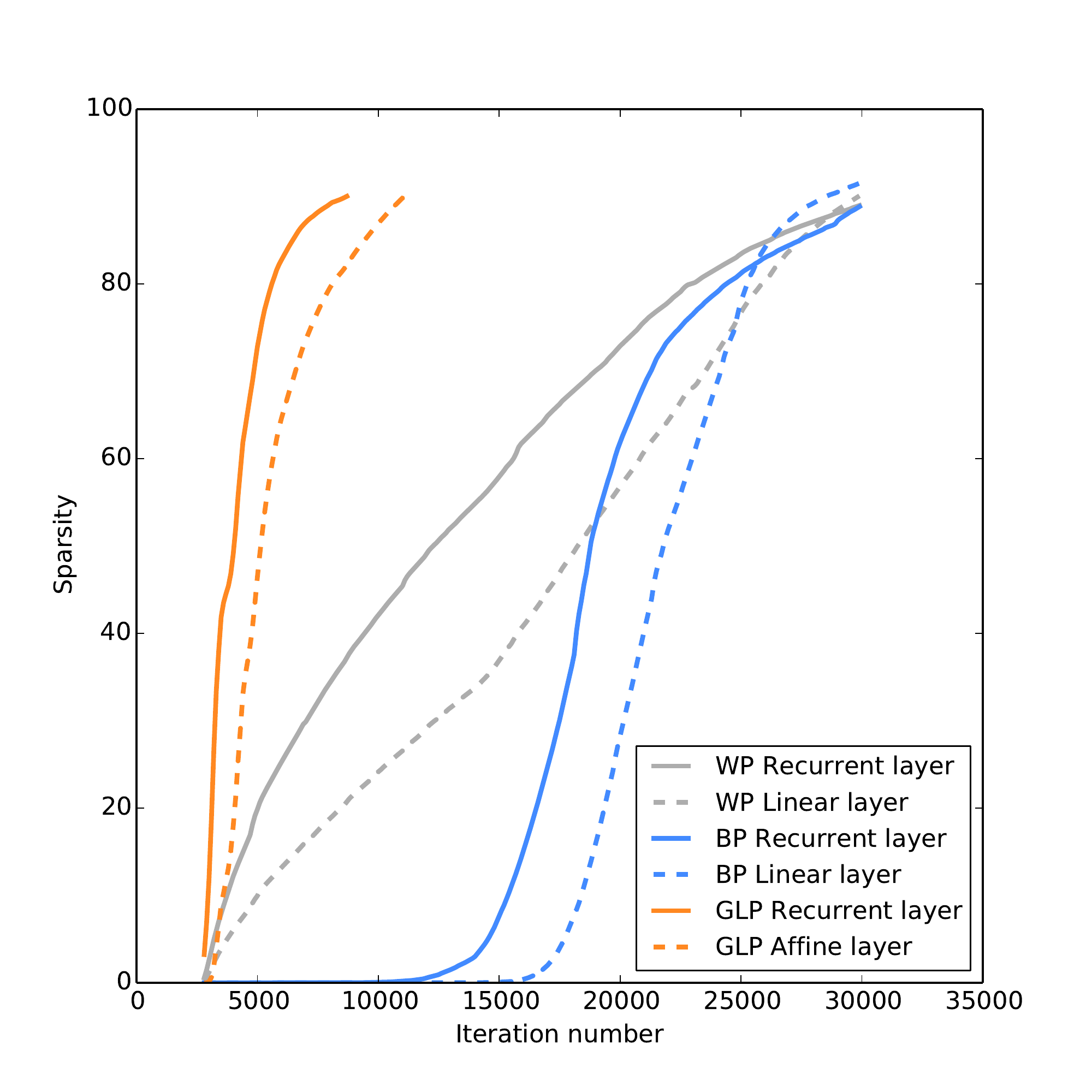} 
\caption{}
\label{fig:prune_sch}
\end{subfigure}
\begin{subfigure}{0.49\linewidth}
\includegraphics[width=0.95\linewidth, height=5cm]{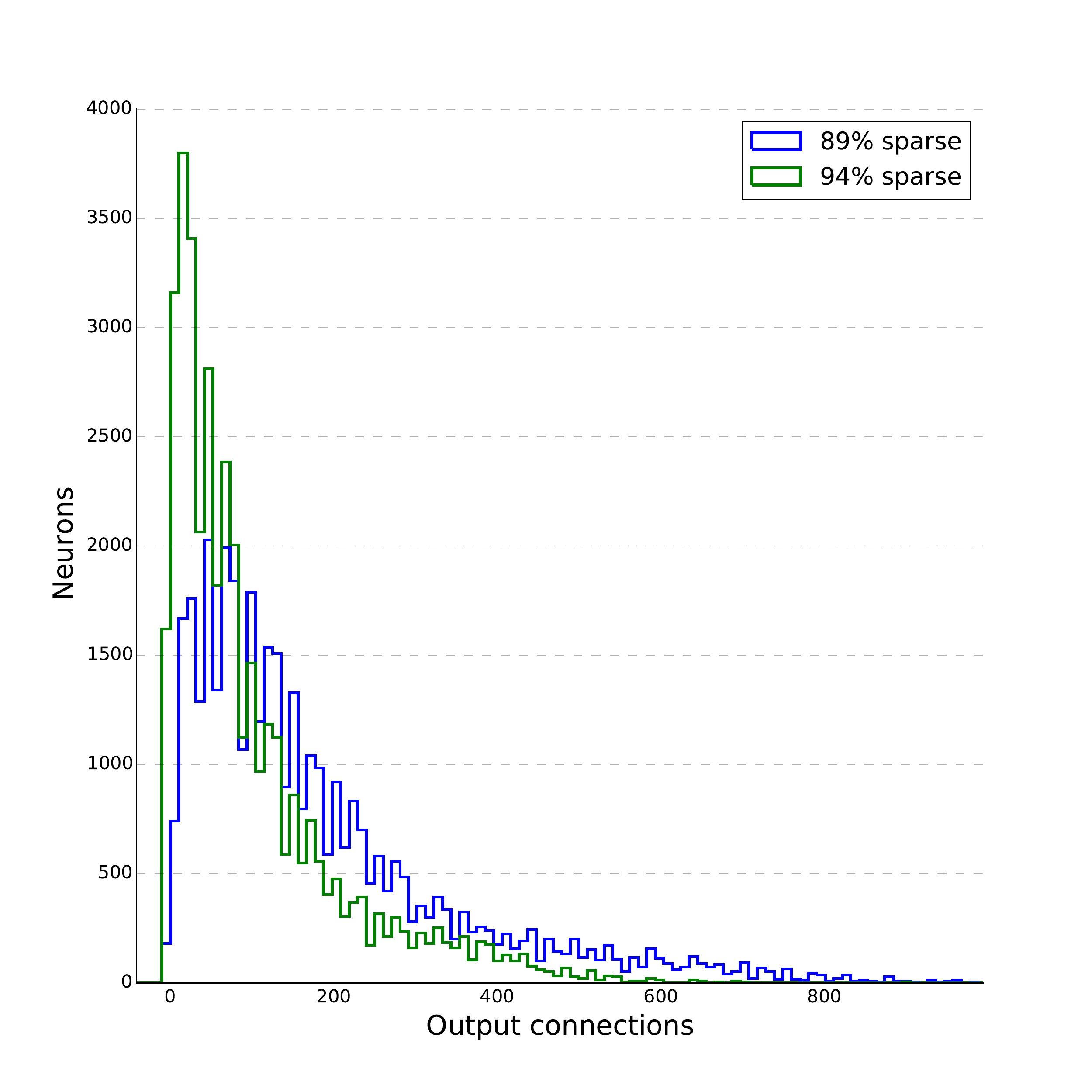}
\caption{}
\label{fig:fan_out_dist}
\end{subfigure}
\label{fig:prune-chars}
\caption{Figure \ref{fig:prune_sch} shows the pruning schedule for two layers in the network for WP, GLP and BP models. The GLP and BP models use block size of 4x4. Figure \ref{fig:fan_out_dist} plots the histogram of the number of output connections for all neurons in the network using block pruning with 4\x4 blocks.}
\end{figure}

In Figure~\ref{fig:prune_sch}, we plot the pruning schedule of a recurrent and linear layer of the bidirectional model trained with BP and Weight Pruning (WP) \citep{narang2017exploring}. For all three algorithms, pruning begins just after the first epoch at 2700 iterations. The BP and GLP models result in a sharper curve with more weights being set to zero in a short span of iterations. In these experiments, we use the \emph{max} function to reduce the blocks to a single value which could be the cause of the sharpness in pruning. Also the GLP model reaches 90\% sparsity just before 10,000 iterations which is significantly earlier than the BP model. GLP training encourages sparsity early on in the training run by pushing the blocks of weights towards zero. 


Figure \ref{fig:fan_out_dist} shows the histogram of the number of output connections for all the neurons in a network for two models with different sparsity pruned with BP. The 94\% sparse model does significantly worse than the 89\% sparse. For the model with 89\% sparsity, only 180 neurons have all their output weights set to zero out of a total of 38270. This model produced good accuracy relative to the dense baseline. However, increasing the sparsity to 94\% for the layer results in 1620 neurons having all zero output weights. Additionally, a lot more neurons have a smaller number of non-zero output weights.


\subsection{Impact of sparsity on accuracy}

Using our baseline RNN model, we run many weight and block pruning experiments, varying hyper-parameters to produce a spectrum of results ranging from 70\% to 97\% sparsity. For these experiments, the models are trained for 20 epochs and the accuracy is measured on the validation set instead of the test set. Therefore, the relative accuracy for these models is slightly different from the results reported in Section \ref{sec:block_sparsity_results}. As shown in Figure \ref{fig:accuracy_sparsity}, models pruned using WP with sparsity less than 95\% have relative accuracy ranging from -20\% to -27\%. Increasing the sparsity for the model beyond 95\% results in 30\% or more accuracy loss. This accuracy "cliff" is earlier for models pruned with block sparsity. For block size 4\x4, models with sparsity greater 90\% yield a relative accuracy loss of 30\% or higher. Similarly, for blocks of 16\x16, models with sparsity greater than 86\% have 30\% or more accuracy loss. A similar trend is observed for block size 32\x32. This indicates that there is a tradeoff between sparsity, block size and accuracy of the model.

\subsection{Sparsity vs Layers}

\begin{figure}
\begin{subfigure}{0.49\linewidth}
\centering
\includegraphics[width=0.95\linewidth, height=5cm]{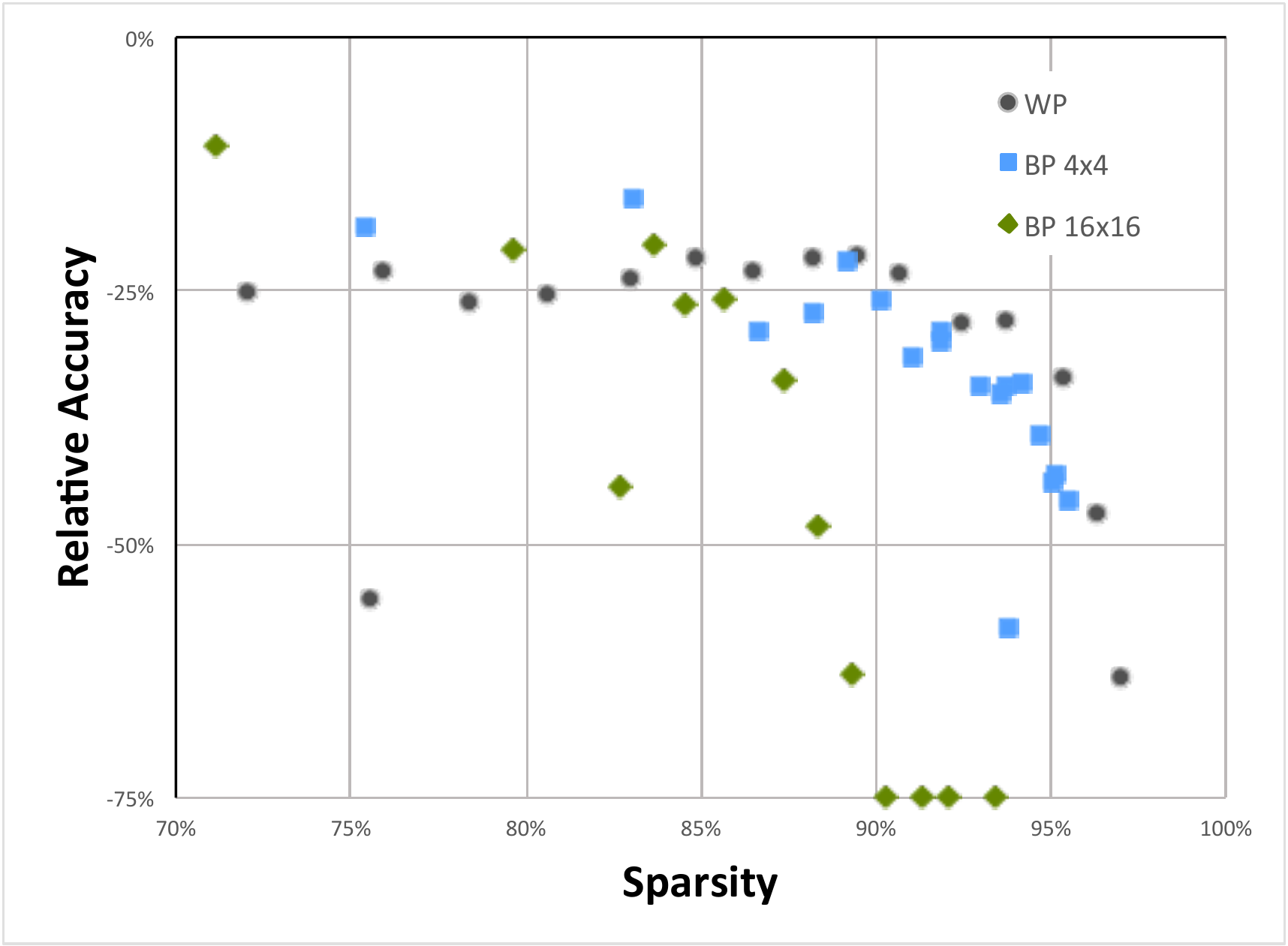} 
\caption{}
\label{fig:accuracy_sparsity}
\end{subfigure}
\begin{subfigure}{0.49\linewidth}
\includegraphics[width=0.95\linewidth, height=5cm]{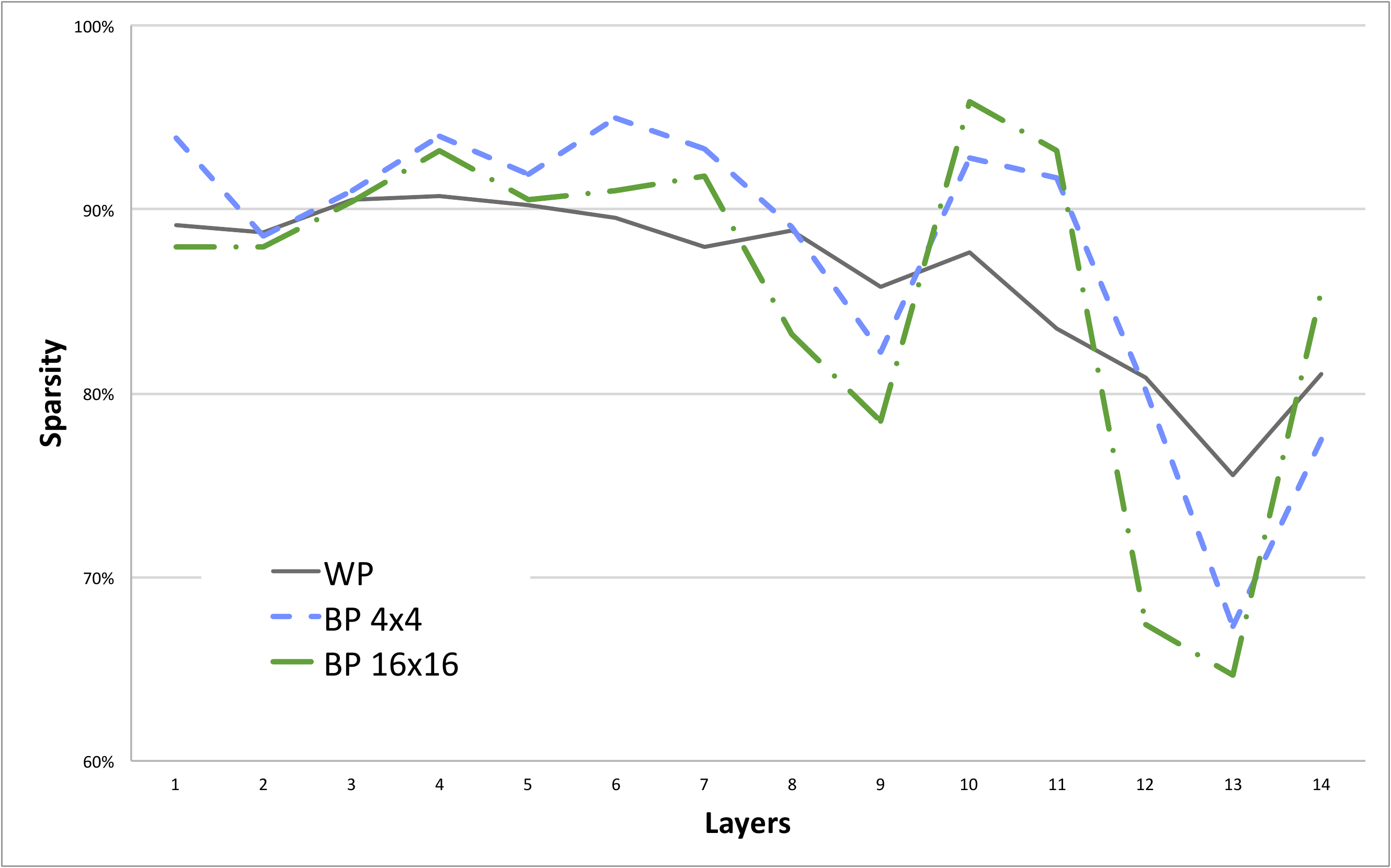}
\caption{}
\label{fig:sparsity_layers}
\end{subfigure}
\label{fig:sparsity-chars}
\caption{Figure \ref{fig:accuracy_sparsity} shows the relative accuracy for different block sizes (4x4, 16x16) and WP for varying sparsity on the RNN 1760 model. Any models with relative accuracy worse than -75\% are capped at 75\%. Figure \ref{fig:sparsity_layers} shows the sparsity of different recurrent layers in the network in the RNN model, pruned using BP and WP.}
\end{figure}

Figure~\ref{fig:sparsity_layers} shows the sparsity of all the recurrent layers in the network using BP and WP. All recurrent layers have the same pruning hyper-parameters. Layer 1 is the first recurrent layer and layer 14 is the final recurrent layer before the CTC cost layer. For both block pruning and weight pruning, we see that the initial layers are pruned more aggressively compared to the final layers. Increasing sparsity in the layers closer to the output results in poor accuracy. Additionally, the variance in sparsity across the layers increases with the block size. This increasing variance makes it harder to increase the block size beyond 32\x32 with the same pruning hyper-parameters for all recurrent layers.



\section{Conclusion and Future Work}
\label{sec:conclusion}

We have demonstrated that using block pruning and group lasso combined with pruning during training we can build block-sparse RNNs that are about as accurate as the dense baseline models. The block-sparse models have significantly fewer parameters than the dense baselines reducing memory requirements. Block-sparse models can take advantage of the underlying hardware efficiently.

We would like to investigate if pruning can be performed even earlier in the training, thereby allowing us to train sparse models. Training sparse models would allow us to reap the benefits of sparsity during training resulting in lesser compute and memory demands. Further work remains to implement efficient block-sparse matrix denese matrix/vector multiplies for GPU and ARM processors that would provide increased speed-up during deployment.

\section*{Acknowledgements}

We would like to thank Song Han, Mohammad Shoeybi, and Markus Kliegl for helpful discussions related to this work. We would also like to thank Varun Arora for creating a figure in the paper.

\newpage
\renewcommand{\bibsection}{\subsubsection*{References}} \small
\bibliographystyle{iclr2018_conference}
\bibliography{bibliography}

\newpage
\appendix \normalsize 

\section{$\ell_1$ and $\ell_{1/2}$ Regularization}
\label{appendix}

Prior to our work with group lasso regularization, we considered $\ell_1$ and $\ell_{1/2}$ regularizers to induce sparsity in the network. These regularizers act on individual weights and could aid in inducing unstructured sparsity in the network. $\ell_1$ regularization is defined as:
$$
L = L_{\text{training}} + \lambda \sum_{i=1}^k |w_i|
$$
where $|w_i|$ is the absolute value of a weight and $k$ is the total number of weights. Note the gradient expression for each weight $w_j$:
$$
\frac{\partial}{\partial w_j} \sum_{i=1}^k |w_i| = sgn(w_j)
$$

As with the group lasso experiments described in \ref{sec:group_lasso}, we explore $\ell_1$ regularization with and without pruning. The weight pruning (WP) algorithm from \citet{narang2017exploring} is used along with regularization. The motivation is the same as group lasso block sparsity experiments: either to guide pruning or to produce sparsity directly.

We also explore $\ell_{1/2}$ regularization which is defined as:
$$
L = L_{\text{training}} + \lambda \sum_{i=1}^k |w_i|^{1/2}
$$

\citet{fan2016} uses $\ell_{1/2}$ regularization to produce sparsity directly. The gradient for $\ell_{1/2}$ regularization is $\frac{1}{2}|w_j|^{-1/2}$. This term is smaller for weights with larger magnitude. Our expectation is that $\ell_{1/2}$ will drive unimportant weights towards zero while leaving large weights relatively unaffected, thus avoiding the accuracy loss associated with excessive regularization. 

For our $\ell_1$ and $\ell_{1/2}$ experiments, we use the Deep Speech 2 Bidirectional RNN baseline model described in Section \ref{sec:experiments}. These models are trained for 25 epochs on our internal training dataset of 2000 hours. The results are reported on a independent test set consisting of 2.9 hours. 

\begin{table}[h]
\caption{$\ell_1$ and $\ell_{1/2}$ results with the bidirectional RNN model with 1760 hidden units}
\label{tab:l1-results}
\begin{center}
 \begin{tabular}{lrrrrl}
\multicolumn{1}{c}{} 
&\multicolumn{1}{c}{\bf \# PARAMS} 
&\multicolumn{1}{c}{} 
&\multicolumn{1}{c}{}
&\multicolumn{1}{c}{\bf RELATIVE} 
&\multicolumn{1}{c}{\bf PRUNING}
\\
\multicolumn{1}{c}{\bf MODEL} 
&\multicolumn{1}{c}{(in millions)} 
&\multicolumn{1}{c}{\bf SPARSITY}
&\multicolumn{1}{c}{\bf CER} 
&\multicolumn{1}{c}{\bf PERF} 
&\multicolumn{1}{c}{\bf ALGORITHM}
\\ \hline \\
RNN Dense & 67   &   0.0\% & 15.36  & 0.0\% & N/A \\
RNN Sparse & 7.3   & 89.2\% & \textbf{17.32}  & -12.8\% & Weight pruning\\
RNN Sparse & 11.2   & 83.6\% & 24.8   & -61.5\% & $\ell_1$\\
RNN Sparse &  7.4   & 89.1\% & \textbf{17.28}  & -12.5\% & $\ell_1$ with pruning  \\
RNN Sparse  &  6.6   & 90.3\% &18.50 & -20.4\% & $\ell_{1/2}$ with pruning \\
\hline
\end{tabular}
\end{center}
\end{table}

Without pruning, $\ell_1$ model results in significantly worse accuracy compared to the dense baseline. Combining $\ell_1$ with weight pruning allows us to recover the loss in accuracy with similar sparsity. The $\ell_{1/2}$ with pruning model performs worse than the $\ell_1$ with pruning model. Comparing the two regularizers, this result indicates that $\ell_1$ is better at guiding pruning than $\ell_{1/2}$, more suitable as a regularizer, or both.

Similar to group lasso experiments, $\ell_1$ regularization experiments require a significantly higher $\lambda$ to achieve high sparsity without any pruning. We suspect that these regularizers would be more successful in inducing sparsity for models that overfit the training training dataset.

\end{document}